\documentclass{article} 
\usepackage{iclr2026_conference,times}

\usepackage{amsmath,amsfonts,bm}

\def\eqref#1{equation~\ref{#1}}

\def\1{\bm{1}}

\DeclareMathAlphabet{\mathsfit}{\encodingdefault}{\sfdefault}{m}{sl}
\SetMathAlphabet{\mathsfit}{bold}{\encodingdefault}{\sfdefault}{bx}{n}

\usepackage{graphicx}
\usepackage{subcaption}
\usepackage{hyperref}
\usepackage{url}
\usepackage{wrapfig}
\usepackage{caption} 
\usepackage[table]{xcolor}   
\usepackage{colortbl}        
\usepackage{amsmath} 
\DeclareMathOperator{\numel}{numel}
\definecolor{HeaderGray}{gray}{0.90}  
\definecolor{RowGray}{gray}{0.96}     
\usepackage{algorithm}
\usepackage{algorithmic}
\usepackage{booktabs}
\title{
\raisebox{-0.4em}{\includegraphics[height=1.4em]{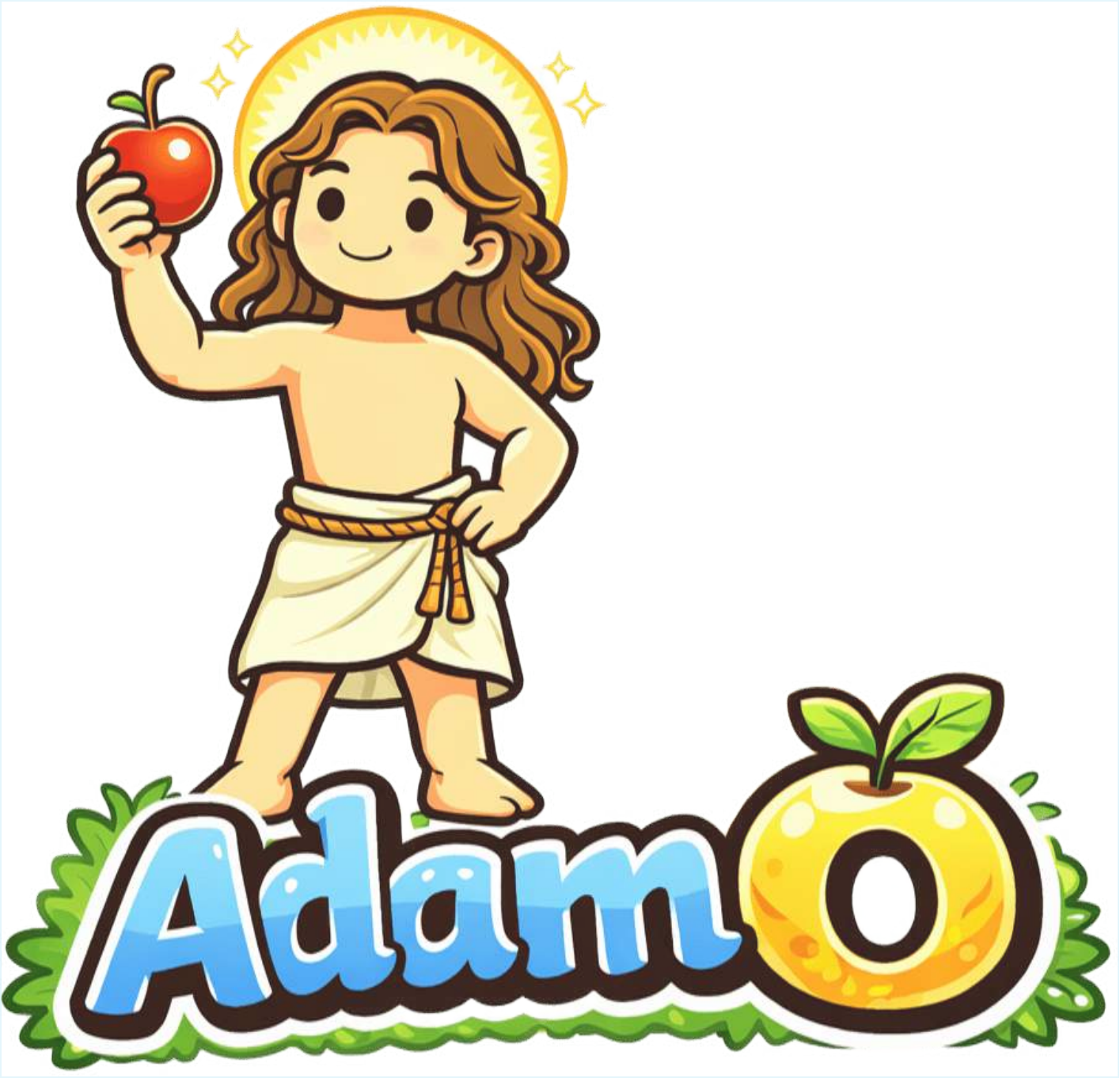}}
      \hspace{0.3em}%
Decoupled Orthogonal Dynamics:\\ Regularization for Deep Network Optimizers}

\iclrfinalcopy

\author{
Hao Chen \\
Beijing University of Posts and Telecommunications \\
\texttt{2022chenhao@bupt.edu.cn}
\And
Jinghui Yuan \\
Northwestern Polytechnical University \\
\texttt{yuanjh@mail.nwpu.edu.cn}
\And
Hanmin Zhang \\
Beijing University of Posts and Telecommunications \\
\texttt{zhanghanmin2024@bupt.edu.cn}
}

\begin{document}

\maketitle
\begin{abstract}
\textbf{Is the standard weight decay in AdamW truly optimal?} Although AdamW decouples weight decay from adaptive gradient scaling, a fundamental conflict remains: \textit{the Radial Tug-of-War}. In deep learning, gradients tend to increase parameter norms to expand effective capacity while steering directions to learn features, whereas weight decay indiscriminately suppresses norm growth. This push--pull interaction induces radial oscillations, injecting noise into Adam's second-moment estimates and potentially degrading delicate tangential feature learning. We argue that magnitude and direction play distinct roles and should be decoupled in optimizer dynamics. We propose \emph{Orthogonal Dynamics Decoupling} and instantiate it as \textbf{AdamO}: an SGD-style update handles the one-dimensional norm control, while Adam's adaptive preconditioning is confined to the tangential subspace. AdamO further incorporates curvature-adaptive radial step sizing and architecture-aware rules and projections for scale-invariant layers and low-dimensional parameters. Experiments on vision and language tasks show that AdamO improves generalization and stability over AdamW without introducing additional complex constraints.
\end{abstract}

\section{Introduction}
Since its inception, AdamW has established itself as a ubiquitous default for training deep neural networks across Computer Vision, Natural Language Processing, and Multimodal Large Language Models~\cite{Self}. Its popularity is largely attributed to decoupling weight decay from adaptive gradient updates. Yet as models scale and tasks grow more demanding, a natural question arises: \textit{Does merely fixing weight decay resolve the underlying geometric conflicts of optimization?}

Recent theoretical critiques suggest the answer is no. \cite{AdamCPR} argue that weight decay is fundamentally a proxy for constraining parameter norms, while standard implementations apply it indiscriminately. \cite{AdamWN} further observes a \textit{bias toward zero} that compels the optimizer to expend computation regrowing weights against the decay force. Notably, these critiques were independently articulated by AdamW’s two original authors, converging on a shared conclusion: the prevailing mechanism is an inefficient compromise that fails to respect the geometry of parameter space.

We attribute this inefficiency to the \textit{Radial Tug-of-War}. During training, a parameter vector plays two distinct roles: its magnitude (norm) governs effective capacity, whereas its direction encodes features. In AdamW~\cite{AdamW} these roles are implicitly entangled: gradients often drive norm growth to expand fitting capability, while weight decay exerts an opposing radial pull; moreover, smaller norms frequently induce larger radial gradients, amplifying oscillations along the radial axis. Because Adam accumulates squared gradients into the variance state $v_t$, such radial noise can inflate variance estimates and contaminate the preconditioner used for delicate tangential updates.

To address this, we propose \emph{Orthogonal Dynamics Decoupling} and instantiate it as \textbf{AdamO}, which strictly separates radial norm control from tangential feature learning. Beyond decoupling, AdamO introduces (i) curvature-adaptive radial step sizing to suppress radial oscillations, and (ii) architecture-aware rules and projections that account for scale-invariant layers and low-dimensional parameters, aligning updates with functionally effective directions. Concretely, we treat radial dynamics as a one-dimensional control problem handled by an SGD-style update with adaptive radial steps, while confining Adam’s adaptive preconditioning to the tangential subspace and applying projections when appropriate. Empirically, this decoupled-and-specialized design—without complex constraints or Lagrange multipliers—consistently outperforms AdamW, highlighting geometric separation as a key ingredient for next-generation optimizers.

\begin{figure}[htbp]
    \centering
    \begin{subfigure}[b]{0.23\textwidth}
        \centering
        \includegraphics[width=\textwidth]{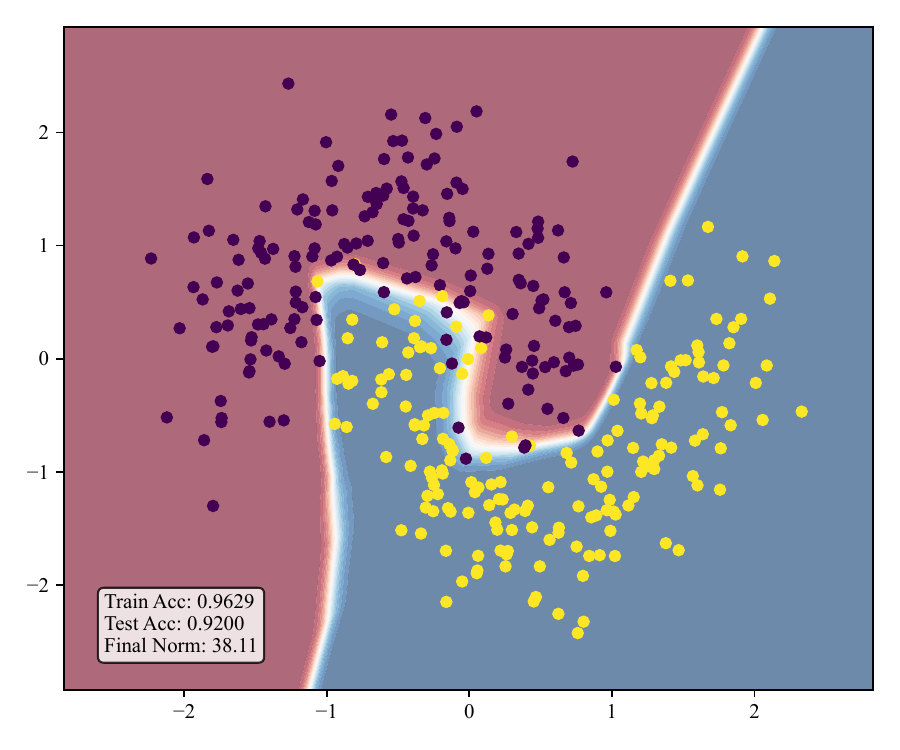}
        \caption{Method: Adam}
        \label{fig:sub1}
    \end{subfigure}
    \hfill
    \begin{subfigure}[b]{0.23\textwidth}
        \centering
        \includegraphics[width=\textwidth]{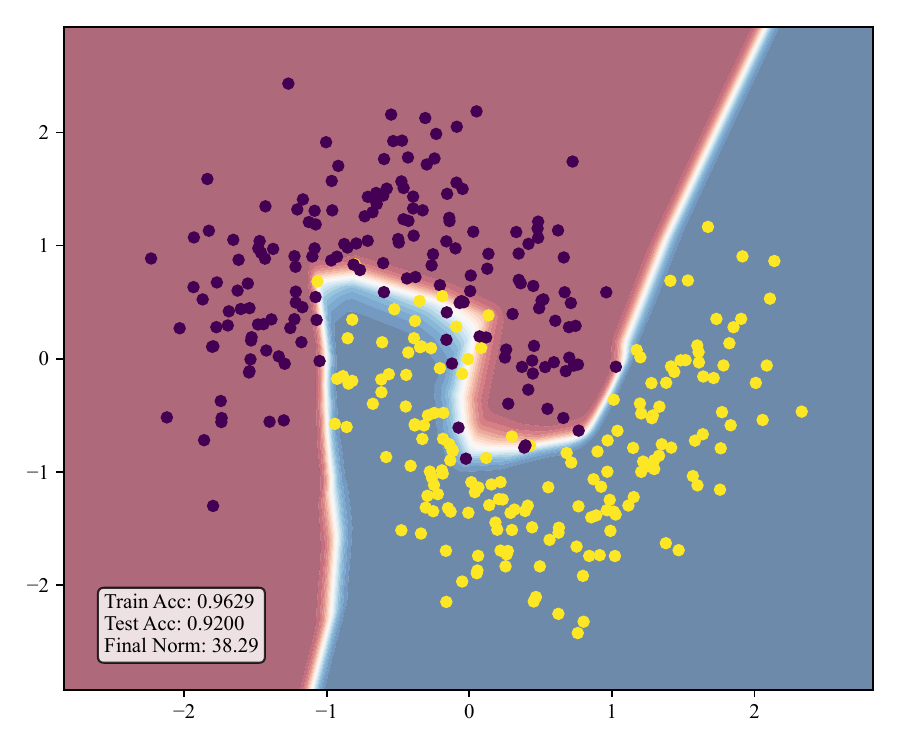}
        \caption{Method: Adam+WD}
        \label{fig:sub2}
    \end{subfigure}
    \hfill
    \begin{subfigure}[b]{0.23\textwidth}
        \centering
        \includegraphics[width=\textwidth]{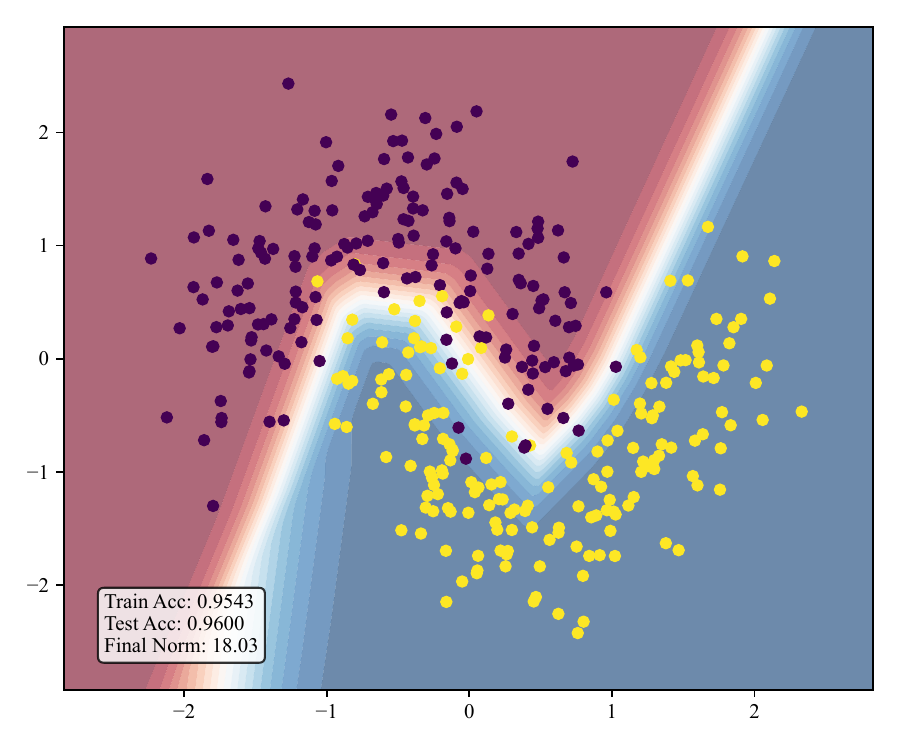}
        \caption{Method: AdamO}
        \label{fig:sub3}
    \end{subfigure}
    \hfill
    \begin{subfigure}[b]{0.23\textwidth}
        \centering
        \includegraphics[width=\textwidth]{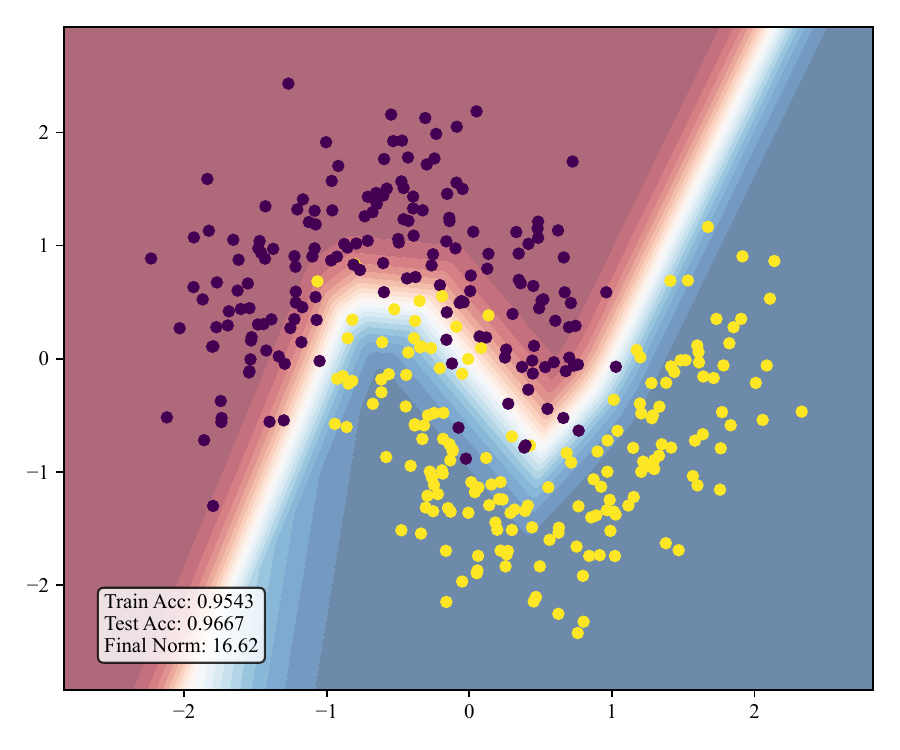}
        \caption{Method:AdamO+WD}
        \label{fig:sub4}
    \end{subfigure}
    \caption{Visualization of neural network training results using Adam and AdamO. AdamO exhibits completely different dynamics compared to Adam, reflected in significantly smaller norms and noticeably smoother decision boundaries.}
    \label{fig:four_subfigs}
\end{figure}

\section{Method}
We propose \textbf{AdamO}, which decouples radial norm control from tangential feature learning and augments this separation with curvature-adaptive radial steps and architecture-aware updates/projections (Algorithm~\ref{algo:adamo}, Appendix~\ref{app:algorithm}).

\subsection{Decoupled Orthogonal Dynamics}
We construct a radial--tangential decomposition per parameter block and enforce subspace closure for \emph{gradients, states, and updates}, yielding strict dynamical decoupling.

\textbf{Radial--tangential projections.}
Let $w\in\mathbb{R}^d$ denote the current parameter vector (tensor blocks are implicitly vectorized) and $\rho=|w|$. For any $z\in\mathbb{R}^d$, define the orthogonal projections w.r.t.\ $w$:
\begin{equation}
\label{eq:proj}
\varphi^{\rho}_{w}(z)
:=\frac{\langle z,w\rangle}{\langle w,w\rangle}\,w,
\qquad
\varphi^{\theta}_{w}(z)
:=z-\varphi^{\rho}_{w}(z),
\end{equation}
so that $z=\varphi^{\rho}_{w}(z)+\varphi^{\theta}_{w}(z)$ with $\varphi^{\rho}_{w}(z)\parallel w$ and $\varphi^{\theta}_{w}(z)\perp w$.
Given a stochastic gradient $g_t=\nabla_w \mathcal{L}_t(w)$ (evaluated at the current iterate), we denote $g_t^\rho=\varphi^\rho_w(g_t)$ and $g_t^\theta=\varphi^\theta_w(g_t)$.

\textbf{Dynamical decoupling via projected states.}
Gradient decomposition alone is insufficient because shared states allow cross-subspace leakage; AdamO maintains \emph{separate} states and re-projects them each step to track the moving subspaces induced by $w$:
\begin{equation}
\label{eq:states}
\begin{aligned}
m_t^\rho &= \beta_1^\rho\,\varphi^\rho_w(m_{t-1}^\rho) + (1-\beta_1^\rho)\,g_t^\rho,\\
m_t^\theta &= \beta_1^\theta\,\varphi^\theta_w(m_{t-1}^\theta) + (1-\beta_1^\theta)\,g_t^\theta,\\
v_t^\theta &= \beta_2^\theta\,v_{t-1}^\theta + (1-\beta_2^\theta)\,(g_t^\theta\odot g_t^\theta),
\end{aligned}
\end{equation}
where $\odot$ denotes elementwise multiplication.
Re-projecting $m_{t-1}^{\rho/\theta}$ is essential because the subspaces rotate with $w$, and re-projection prevents state interference.

\textbf{Pure radial weight decay.}
Unlike isotropic decay (as in AdamW), AdamO applies $L_2$ regularization \emph{purely radially}:
\begin{equation}
\label{eq:radial_wd}
w^{\text{decay}} \;=\; (1-\eta_{\rho,t}\lambda)\,w,
\end{equation}
where $\lambda$ is the decay coefficient and $\eta_{\rho,t}$ is the (possibly time-varying) radial step size.
This scales $|w|$ without changing $\theta=w/|w|$, avoiding directional contamination.

\textbf{Subspace-wise updates.}
We treat norm control as a 1D problem updated by an SGD-style radial step, while confining Adam’s adaptive preconditioning to the tangential subspace.
With bias corrections $\hat m_t^\rho=m_t^\rho/(1-(\beta_1^\rho)^t)$, $\hat m_t^\theta=m_t^\theta/(1-(\beta_1^\theta)^t)$ and $\hat v_t^\theta=v_t^\theta/(1-(\beta_2^\theta)^t)$, we compute
\begin{equation}
\label{eq:update_components}
\Delta w_t^\rho=\eta_{\rho,t}\,\varphi^\rho_w(\hat m_t^\rho),\qquad
\Delta w_t^\theta=\eta_\theta\,\varphi^\theta_w\!\left(\frac{\hat m_t^\theta}{\sqrt{\hat v_t^\theta}+\epsilon}\right),
\end{equation}
and update $w^+=w^{\text{decay}}-(\Delta w_t^\rho+\Delta w_t^\theta)$.
Even after preconditioning, we explicitly apply $\varphi^\theta_w(\cdot)$ to ensure $\Delta w_t^\theta\perp w$, preserving subspace closure at the level of \emph{gradients, states, and updates}.

\subsection{Curvature-Adaptive Radial Step Size}
AdamO adapts \emph{only} the radial step size using a lightweight curvature proxy, slowing down in high-curvature regions and speeding up on flatter ones.

\textbf{Curvature proxy with exponential smoothing.}
We estimate curvature by the squared change in stochastic gradients and smooth it with an exponential moving average:
\begin{equation}
\label{eq:curv}
\kappa_t := \|g_t-g_{t-1}\|^2,\qquad
\tau_t := \beta_\tau \tau_{t-1} + (1-\beta_\tau)\kappa_t,
\end{equation}
where $\tau_t$ tracks a stable curvature scale and $\beta_\tau\in[0,1)$ controls smoothing.

\textbf{Adaptive radial learning rate.}
Given a target scale $\tau_{\text{target}}$, we set
\begin{equation}
\label{eq:lr_adapt}
\eta_{\rho,t}
\;:=\;
\frac{\eta_{\rho}}{\sqrt{\tau_t/\tau_{\text{target}}+\epsilon}}.
\end{equation}
This simple normalization yields robust behavior across training phases without altering tangential Adam preconditioning, keeping the decoupling principle intact.

\subsection{Architecture-Aware Updates and Projections}
AdamO is parameter-aware: it uses a simplified Adam update for low-dimensional parameters and an AdamP-style tangential-only rule for scale-invariant layers, avoiding uninformative radial steps.

\textbf{Dimension-aware fast path for low-dimensional parameters.}
For effectively low-dimensional parameters (e.g., biases and norm affine terms), when $\dim(w)\le 1$ or $\numel(w)<d_{\text{th}}$ we apply a standard Adam update
\begin{equation}
\label{eq:lowdim}
w^+ \;=\; w - \alpha\,\eta_\theta \frac{\hat m_t}{\sqrt{\hat v_t}+\epsilon},
\end{equation}
where $(m_t,v_t)$ are the usual Adam moments for this block and $\alpha\in(0,1]$ is a stabilization factor.

\textbf{Projection for scale-invariant layers (tangential-only update).}
For scale-invariant layers (e.g., BatchNorm/LayerNorm), radial steps are largely uninformative; we therefore apply the tangential step only:
\begin{equation}
\label{eq:scale_invariant}
\Delta w_t \leftarrow \Delta w_t^\theta,
\end{equation}
which can be viewed as an AdamP-style projection constraint expressed naturally within our decoupled framework.

\section{Experiments}

We evaluate AdamO on image classification (CIFAR-100~\cite{Krizhevsky2009LearningML}) and modular arithmetic Grokking~\cite{power2022grokkinggeneralizationoverfittingsmall}; due to space, we focus on CIFAR-100 in the main text and report Grokking in Appendix~\ref{app:grokking}.

\subsection{Experimental Setup}
Following the protocol of~\cite{AdamCPR} (see Appendix~\ref{app:dataset} for task/model details), we compare Adam~\cite{Adam}, AdamW~\cite{AdamW}, AdamP~\cite{AdamP}, and AdamO variants under the same training budget and scheduler (baseline notes in Appendix~\ref{app:baselines}). All runs are conducted on a single \textbf{NVIDIA GeForce RTX 3090}; each setting is repeated three times and we report mean $\pm$ standard deviation.

Unless otherwise specified, CIFAR-100 is trained for 300 epochs with batch size 128. We use MultiStepLR with milestones $\{50,100,150,200,250\}$ and $\gamma=0.2$, and a 10-epoch warmup (initial LR $=0.1\times\eta_\theta$). For AdamO, we set tangential LR $\eta_\theta=8\times10^{-4}$, radial LR $\eta_\rho=5\times10^{-3}$, and pure-radial weight decay $\lambda=2\times10^{-4}$, with Adam defaults $\beta_1=0.9,\beta_2=0.999$. From epoch 200 onward, we enable SWA (LR $10^{-4}$) and label smoothing (0.1), and activate projection-related settings ($\delta=0.1$, $\texttt{wd\_ratio}=0.5$).

\subsection{Main Results and Ablations}
Table~\ref{tab:comprehensive_results} reports CIFAR-100 results and key ablations. AdamO reaches \textbf{79.74$\pm$0.09}\% accuracy, improving over AdamW by \textbf{+4.99} points (79.74 vs 74.75), whereas AdamP provides only a minor gain (75.07 vs 74.75), suggesting that projection alone does not resolve the dominant instability. Ablations show that removing curvature-adaptive radial stepping drops accuracy to 75.21, and disabling dimension-aware handling or projection reduces it to 75.99 and 76.17, respectively. Finally, AdamO-Isotropic is statistically indistinguishable from AdamW (74.82 vs 74.75), reinforcing that \emph{radial-only regularization} is essential: orthogonal decomposition without it yields little benefit.

\begin{table*}[t]
\centering
\caption{CIFAR-100 accuracy (\%). Here AdamO-Isotropic uses isotropic decay.}
\label{tab:comprehensive_results}

\setlength{\tabcolsep}{10pt}
\renewcommand{\arraystretch}{1.15}
\rowcolors{2}{RowGray}{white}

\begin{tabular}{l c l c}
\toprule
\rowcolor{HeaderGray}
\textbf{Configuration} & \textbf{Acc (\%)} & \textbf{Configuration} & \textbf{Acc (\%)} \\
\midrule
Adam & 74.48 $\pm$ 0.12 & AdamW & 74.75 $\pm$ 0.15 \\
AdamP & 75.07 $\pm$ 0.18 & AdamO-Isotropic & 74.82 $\pm$ 0.12 \\
AdamO (w/o Projection) & 76.17 $\pm$ 0.14 & AdamO (w/o Dimension) & 75.99 $\pm$ 0.15 \\
AdamO (w/o Curvature) & 75.21 $\pm$ 0.11 & \textbf{AdamO} & \textbf{79.74 $\pm$ 0.09} \\
\bottomrule
\end{tabular}

\rowcolors{2}{}{} 
\end{table*}

\begin{figure*}[t]
\centering
{\renewcommand\thesubfigure{\roman{subfigure}}

\begin{subfigure}[t]{0.62\textwidth}
  \centering
  \includegraphics[width=\linewidth]{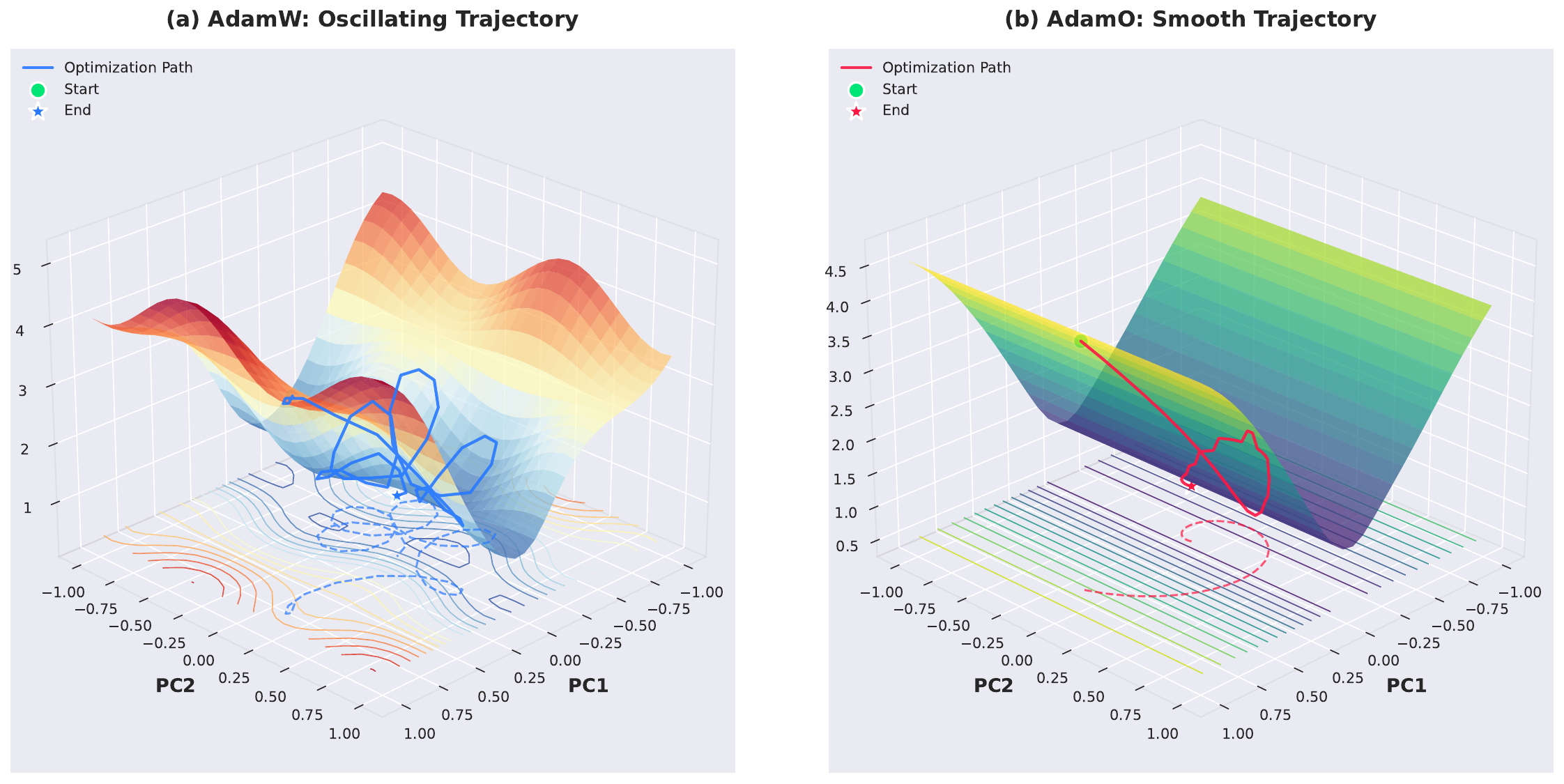}
  \caption{Optimization dynamics visualization.}
  \label{fig:loss_landscape}
\end{subfigure}\hfill
\begin{subfigure}[t]{0.36\textwidth}
  \centering
  \includegraphics[width=\linewidth]{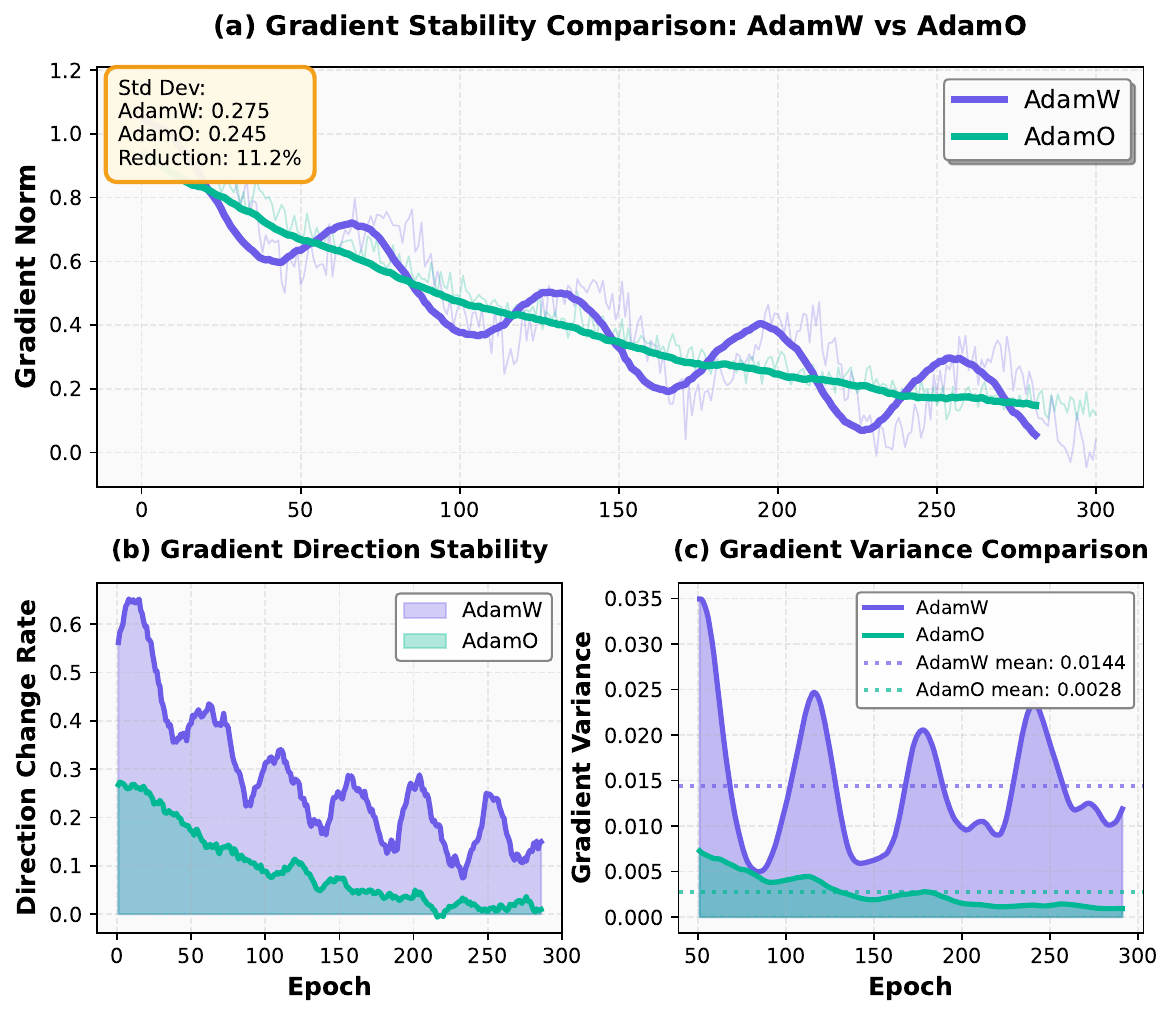}
  \caption{Gradient stability.}
  \label{fig:gradient_stability}
\end{subfigure}

}
\end{figure*}

\subsection{Training Dynamics}
We visualize the optimization dynamics (Fig.~\ref{fig:loss_landscape}) following the 2D subspace method of~\cite{landscape}. AdamW exhibits noticeably stronger radial wandering across contour level sets, whereas AdamO follows a smoother, more directed trajectory. The gradient statistics and the evolution of parameter norms (Fig.~\ref{fig:gradient_stability}) indicate that AdamO substantially reduces gradient-norm fluctuations, yielding a smoother trajectory with a smaller parameter norm.

\section{Conclusion}
We propose \textbf{AdamO}, which strictly decouples radial norm control from tangential feature learning in optimizer dynamics, and further aligns updates with functionally effective directions via curvature-adaptive radial step sizing and architecture-aware rules and projections. Across vision and language tasks, AdamO consistently outperforms AdamW/AdamP, yielding more stable training dynamics and improved generalization without introducing additional complex constraints. We hope this work advances the paradigm of \emph{subspace-specialized} optimization and provides a simple yet effective design principle for next-generation adaptive optimizers.

\bibliography{iclr2026_conference}
\bibliographystyle{iclr2026_conference}

\appendix
\section{Appendix}

\subsection{Full AdamO Pseudocode}
\label{app:algorithm}
We provide the full AdamO pseudocode~\ref{algo:adamo} for reproducibility; the main text focuses on the geometric formulation and the resulting update rules.

\begin{small}
\begin{algorithm}[t]
\caption{\emph{AdamO}: fully decoupled orthogonal dynamics with curvature-adaptive radial step sizing and architecture-aware updates/projections.}
\label{algo:adamo}
\begin{algorithmic}[1]
\REQUIRE $\eta_\theta,\eta_\rho$ (base tangential/radial learning rates), $\lambda$ (pure-radial weight decay), $\epsilon$
\REQUIRE $\beta_1^\theta,\beta_2^\theta,\beta_1^\rho\in[0,1)$ (EMA rates), $\beta_\tau\in[0,1)$ (curvature smoothing)
\REQUIRE $\tau_{\text{target}}$ (target curvature scale), $d_{\text{th}}$ (low-dim threshold), $\alpha\in(0,1]$
\REQUIRE \textsc{LowDim}$(w)$, \textsc{ScaleInv}$(w)$ predicates; projection operators $\varphi_w^\rho(\cdot),\varphi_w^\theta(\cdot)$

\STATE Initialize $t\!\gets\!0$; $m^\theta\!\gets\!0,\ v^\theta\!\gets\!0,\ m^\rho\!\gets\!0$; $\tau\!\gets\!\tau_{\text{target}}$; $g^{-}\!\gets\!0$
\WHILE{not converged}
  \STATE $t\!\gets\!t+1$; $g\!\gets\!\nabla_w \mathcal{L}_t(w)$
  \STATE $\kappa\!\gets\!\|g-g^{-}\|^2$; $\tau\!\gets\!\beta_\tau\tau+(1-\beta_\tau)\kappa$; $g^{-}\!\gets\!g$
  \STATE $\eta_{\rho,t}\!\gets\!\eta_\rho/\sqrt{\tau/\tau_{\text{target}}+\epsilon}$

  \IF{\textsc{LowDim}$(w)$}
    \STATE $m^\theta\!\gets\!\beta_1^\theta m^\theta+(1-\beta_1^\theta)g$;\ $v^\theta\!\gets\!\beta_2^\theta v^\theta+(1-\beta_2^\theta)(g\odot g)$
    \STATE $\hat m^\theta\!\gets\!m^\theta/(1-(\beta_1^\theta)^t)$;\ $\hat v^\theta\!\gets\!v^\theta/(1-(\beta_2^\theta)^t)$
    \STATE $w\!\gets\!w-\alpha\,\eta_\theta\,\hat m^\theta/(\sqrt{\hat v^\theta}+\epsilon)$; \textbf{continue}
  \ENDIF

  \STATE $g^\rho\!\gets\!\varphi_w^\rho(g)$;\ $g^\theta\!\gets\!g-g^\rho$
  \STATE $m^\rho\!\gets\!\beta_1^\rho\,\varphi_w^\rho(m^\rho)+(1-\beta_1^\rho)g^\rho$
  \STATE $m^\theta\!\gets\!\beta_1^\theta\,\varphi_w^\theta(m^\theta)+(1-\beta_1^\theta)g^\theta$
  \STATE $v^\theta\!\gets\!\beta_2^\theta v^\theta+(1-\beta_2^\theta)(g^\theta\odot g^\theta)$
  \STATE $\hat m^\rho\!\gets\!m^\rho/(1-(\beta_1^\rho)^t)$;\ $\hat m^\theta\!\gets\!m^\theta/(1-(\beta_1^\theta)^t)$;\ $\hat v^\theta\!\gets\!v^\theta/(1-(\beta_2^\theta)^t)$

  \STATE $\Delta^\rho\!\gets\!\eta_{\rho,t}\,\varphi_w^\rho(\hat m^\rho)$
  \STATE $\Delta^\theta\!\gets\!\eta_\theta\,\varphi_w^\theta\!\big(\hat m^\theta/(\sqrt{\hat v^\theta}+\epsilon)\big)$
  \STATE $\Delta\!\gets\!\Delta^\theta$ \textbf{if} \textsc{ScaleInv}$(w)$ \textbf{else} $\Delta^\rho+\Delta^\theta$
  \STATE $w\!\gets\!(1-\eta_{\rho,t}\lambda)\,w-\Delta$
\ENDWHILE
\RETURN $w$
\end{algorithmic}
\end{algorithm}
\end{small}

\subsection{Datasets, Models, and Training Protocol}
\label{app:dataset}
We summarize the CIFAR-100 setup, model choices, and training protocol, and explain why BatchNorm-equipped architectures are informative for evaluating scale-invariance and projection behavior.

\paragraph{CIFAR-100.}
CIFAR-100 contains 100 classes with 50k training images and 10k test images at $32\times32$ resolution. We use standard data augmentation: random crop with padding 4 and random horizontal flip.

\paragraph{Model architecture.}
We use ResNet-18 with BatchNorm as the primary backbone. BatchNorm introduces scale-invariant components where radial perturbations can be functionally uninformative, making the setting particularly suitable for stress-testing architecture-aware projections under AdamO.

\paragraph{Training protocol (CIFAR-100).}
Unless otherwise noted, we train for 300 epochs with batch size 128, using MultiStepLR with milestones $\{50,100,150,200,250\}$ and $\gamma=0.2$, plus a 10-epoch warmup. From epoch 200 onward, we enable SWA and label smoothing, matching the main-text protocol for all optimizers to isolate optimizer effects.

\subsection{Baseline Notes}
\label{app:baselines}
We briefly summarize the baselines and emphasize that all optimizers are compared under the same compute budget and scheduler to isolate optimizer-induced effects.

\paragraph{Adam / AdamW / AdamP.}
Adam is the standard adaptive first-order optimizer. AdamW decouples weight decay from adaptive scaling. AdamP introduces a projection heuristic motivated by scale-invariant weights to suppress ineffective updates; we keep all non-optimizer training choices identical across methods.

\subsection{Additional CIFAR-100 Diagnostics}
\label{app:diagnostics}
We include two auxiliary diagnostics that directly support the main-text findings: (i) validation-accuracy trajectories over the first 200 epochs on CIFAR-100, and (ii) a 2D hyperparameter sensitivity heatmap comparison between AdamW and AdamO (Appendix~\ref{app:hparam}).

\begin{figure}[t]
  \centering
  \includegraphics[width=\linewidth]{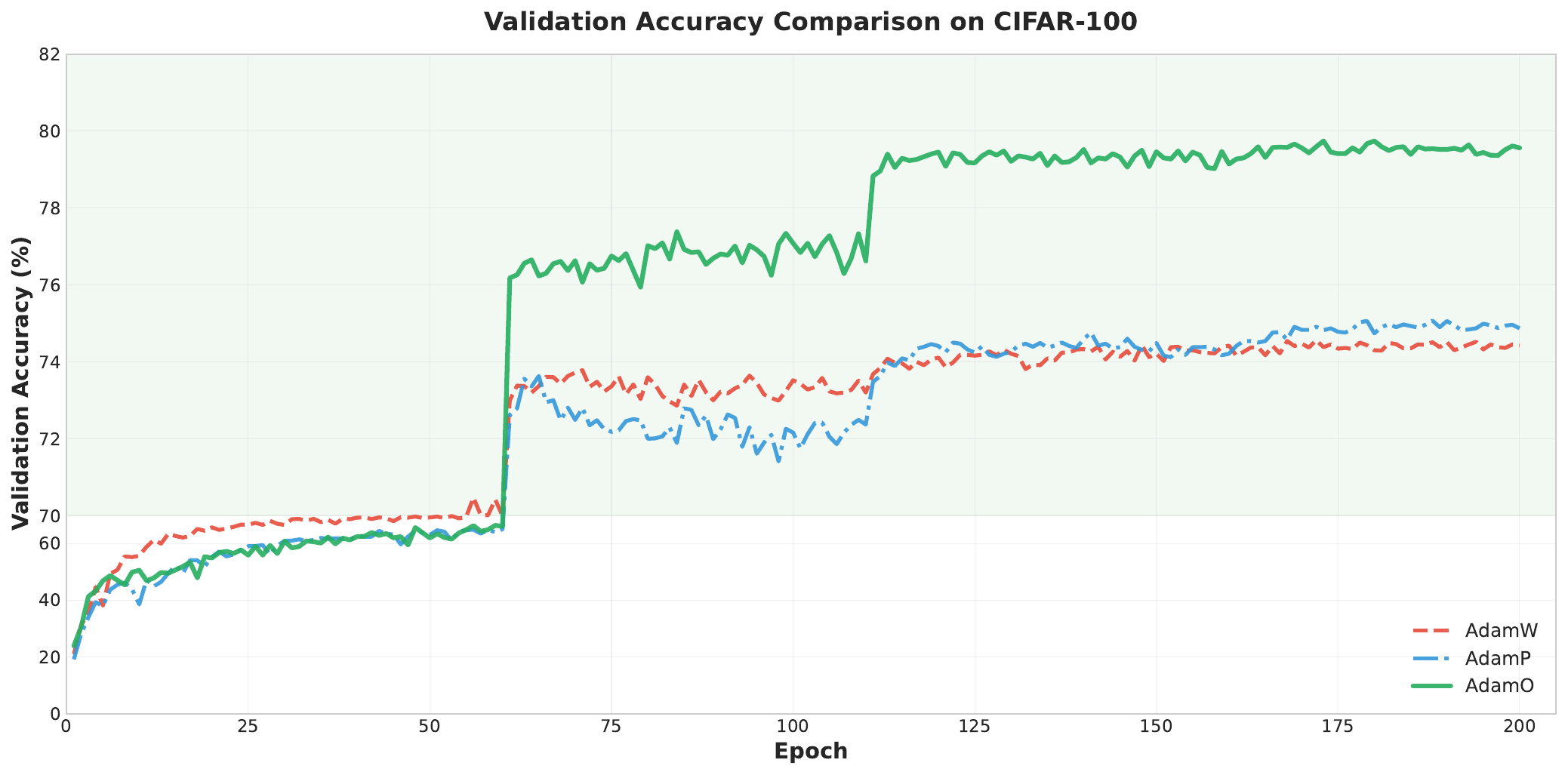}
  \caption{Validation accuracy over 200 epochs on CIFAR-100 for AdamW, AdamP, and AdamO under the same training budget and scheduler. AdamO consistently attains higher validation accuracy and shows larger gains after learning-rate drops.}
  \label{fig:valid_acc_200e}
\end{figure}

\subsection{Hyperparameter Sensitivity}
\label{app:hparam}
We evaluate hyperparameter sensitivity via 2D grid search and visualize validation accuracy as heatmaps. For AdamW, we sweep the standard pair \emph{(learning rate, weight decay)}; for AdamO, we sweep \emph{(tangential learning rate $\eta_\theta$, radial learning rate $\eta_\rho$)}. Brighter cells indicate higher accuracy.

Across the grid, AdamO exhibits a broader and more contiguous high-accuracy region, whereas AdamW’s best-performing region is more localized, indicating higher sensitivity. This suggests that AdamO reduces tuning burden and improves robustness to hyperparameter choices.

\begin{figure}[t]
  \centering
  \includegraphics[width=\linewidth]{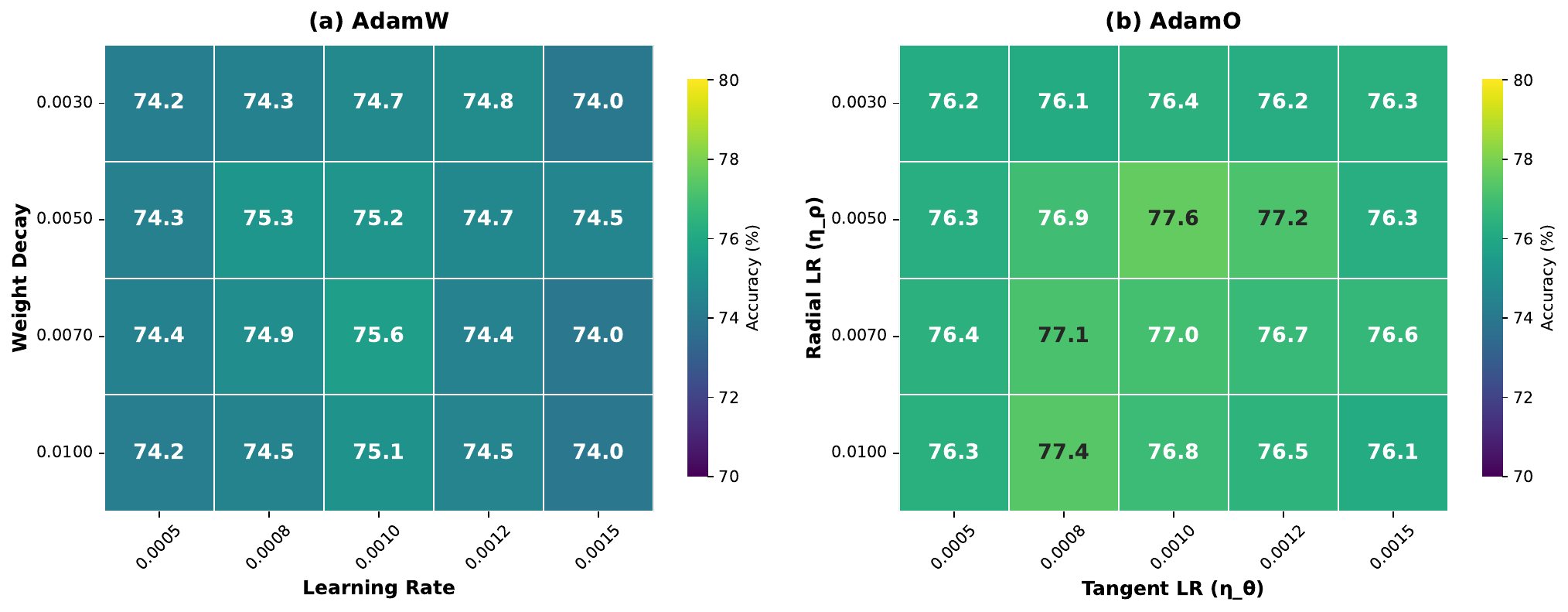}
  \caption{Hyperparameter sensitivity heatmaps on CIFAR-100. \textbf{Left:} AdamW grid over (learning rate, weight decay). \textbf{Right:} AdamO grid over (tangential LR $\eta_\theta$, radial LR $\eta_\rho$). AdamO maintains strong performance across a wider region.}
  \label{fig:hyperparameter_sensitivity}
\end{figure}

\subsection{Grokking Setup and Results}
\label{app:grokking}
We evaluate AdamO on modular-arithmetic Grokking under strong regularization to further probe its regularization behavior. We report final performance and a weight-decay ablation in tables (no curve figure is included in the current draft).

\paragraph{Task and split.}
We consider modular addition $(a+b)\bmod p$ with $p=97$. We train on 30\% of the pairs and evaluate on the remaining 70\%, a regime known to induce the characteristic grokking phase transition under sufficiently strong weight decay.

\paragraph{Model and optimization.}
We use a 2-layer MLP with hidden width 128 and ReLU activations. We train for 5000 epochs with batch size 512, learning rate $10^{-3}$, and weight decay $1.0$, without learning-rate schedules or data augmentation.

\paragraph{Metrics.}
In addition to final test accuracy, we report the \emph{grokking epoch}, defined as the first epoch at which test accuracy exceeds 95\%, to characterize when the phase transition occurs.

\begin{table}[t]
\centering
\caption{Grokking performance on modular addition ($p=97$).}
\label{tab:grokking_results}
\begin{tabular}{lcc}
\toprule
Optimizer & Test Acc (\%) & Grokking Epoch \\
\midrule
Adam      & 89.27 & -- \\
AdamW     & 99.02 & 2508 \\
AdamP     & 99.00 & 2425 \\
AdamO     & \textbf{99.13} & 2785 \\
\bottomrule
\end{tabular}
\end{table}

AdamO achieves the best final accuracy while exhibiting a later transition, consistent with the interpretation that stricter capacity control can delay the memorization-to-generalization phase change yet yield a more robust final solution.

\begin{table}[t]
\centering
\caption{Grokking ablation on weight-decay mechanisms.}
\label{tab:grokking_ablation}
\begin{tabular}{lcc}
\toprule
Configuration & Decay Direction & Test Acc (\%) \\
\midrule
AdamW           & isotropic & 99.02 \\
AdamO-Isotropic & isotropic (within AdamO) & 98.95 \\
AdamO           & radial-only & \textbf{99.13} \\
\bottomrule
\end{tabular}
\end{table}

The ablation reinforces that, under strong regularization, orthogonal decomposition alone is not sufficient: isotropic decay can over-regularize feature-encoding directions, whereas radial-only decay better isolates capacity control and preserves tangential learning.

\end{document}